\documentclass[10pt,twocolumn,letterpaper]{article}

\usepackage{iccv}
\usepackage{times}
\usepackage{epsfig}
\usepackage{graphicx}
\usepackage{amsmath}
\usepackage{amssymb}
\usepackage{bm}
\usepackage{enumitem,kantlipsum}
\usepackage{breqn}
\usepackage{xr}
\usepackage[toc,page]{appendix}
\usepackage{booktabs}
\usepackage[T1]{fontenc}
\usepackage[pagebackref=true,breaklinks=true,letterpaper=true,colorlinks,bookmarks=false]{hyperref}

 \iccvfinalcopy 

\def\iccvPaperID{4951} 

\ificcvfinal\fi
\begin{document}

\title{Evaluating Bayesian Deep Learning Methods for Semantic Segmentation}

\author{Jishnu Mukhoti\\
University of Oxford\\
{\tt\small jishnu.mukhoti@cs.ox.ac.uk}
\and
Yarin Gal\\
University of Oxford\\
{\tt\small yarin@cs.ox.ac.uk}
}

\maketitle

\begin{abstract}
Deep learning has been revolutionary for computer vision and semantic segmentation in particular, with Bayesian Deep Learning (BDL) used to obtain uncertainty maps from deep models when predicting semantic classes. This information is critical when using semantic segmentation for autonomous driving for example. 
Standard semantic segmentation systems have well-established evaluation metrics.
However, with BDL's rising popularity in computer vision we require new metrics to evaluate whether a BDL method produces better uncertainty estimates than another method. In this work we propose three such metrics to evaluate BDL models designed specifically for the task of semantic segmentation. We modify DeepLab-v3+, one of the state-of-the-art deep neural networks, and create its Bayesian counterpart using MC dropout and Concrete dropout as inference techniques. We then compare and test these two inference techniques on the well-known Cityscapes dataset using our suggested metrics. Our results provide new benchmarks for researchers to compare and evaluate their improved uncertainty quantification in pursuit of safer semantic segmentation.
\end{abstract}

\section{Introduction} \label{sec:intro}

Deep learning has had tremendous success in quite a few fields including computer vision \cite{cv1, cv2, cv3, cv4}, natural language processing \cite{nlp1, nlp2, nlp3, nlp4, nlp5}, speech recognition \cite{speech1, speech2, speech3, speech4, speech5}, bioinformatics \cite{bio1, bio2, bio3} and others. However, most deep learning models produce point-estimates as outputs and hence we do not gain any knowledge about the confidence of the model in its predictions. With the increasing use of AI systems in real-life scenarios like autonomous driving \cite{autonomousdriving1, autonomousdriving2, autonomousdriving3, autonomousdriving4, autonomousdriving5} and medical diagnosis \cite{medical1, medical2, medical3, medical4}, there are many cases in which additional knowledge about the model's confidence, i.e.\ capturing whether the model is essentially `guessing at random', becomes not only useful but essential \cite{aisafety}.

The development of new computationally light-weight, scalable methods of performing approximate Bayesian inference in deep neural networks has enabled these models to estimate their uncertainty in addition to making predictions \cite{galdropout, concretedropout, deepensembles, bayesbybackprop, vmg, dgp}. However, as we do not have ground truth uncertainties, we cannot use the conventional methods of evaluation to compare and benchmark these models.  Furthermore, the metrics designed for evaluating model performance are often task dependent. For instance, the intersection-over-union (IOU) \cite{fcn} metric is heavily used in computer vision problems like object detection and semantic segmentation. Extending on these ideas, in this work, we propose new specialised metrics to evaluate Bayesian models designed for the task of semantic segmentation.

\begin{figure*}[t]
\begin{center}
   \includegraphics[width=\textwidth]{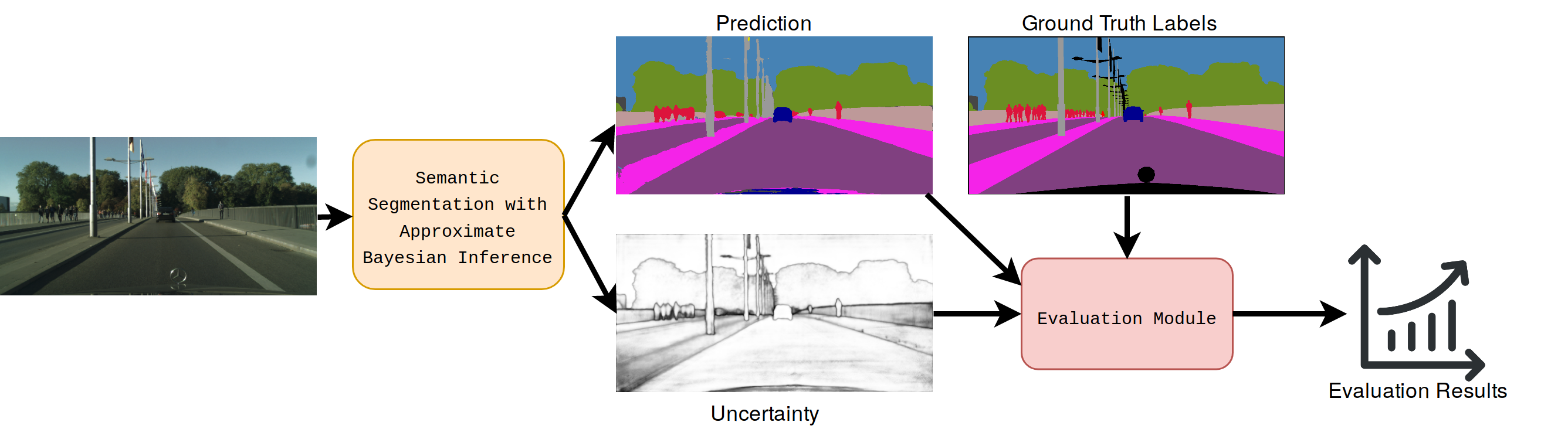}
\end{center}
\vspace{-1\baselineskip}
   \caption{High level overview of the system proposed in this work. The input is first passed through a Bayesian neural network which produces pixel-wise predictions as well as pixel-wise uncertainty estimates. The ground truth labels, predictions and uncertainties are then sent to the performance evaluation module which returns the values of the metrics designed for evaluating the model.}
\vspace{-1\baselineskip}
\label{fig:overview}
\end{figure*}

Semantic segmentation \cite{fcn} is a difficult problem in computer vision which requires pixel-level understanding of an image. A Bayesian model for semantic segmentation will not only produce predictions for each pixel but also generate pixel-wise uncertainty estimates. As mentioned above, evaluating such a Bayesian model is a challenging task because unlike predictions, we do not have a strong definition of what a good uncertainty estimate is. Hence, we have to judge the quality of the model uncertainty based on how accurate the model is for the same input. We require metrics which look at both the model predictions and uncertainties and take into account general desiderata about when a model should be uncertain about its predictions. In particular, we use the following two intuitive desiderata:
\vspace{-1.5mm}
\begin{enumerate}[leftmargin=*]
\itemsep-0.3em
\item \textbf{Desideratum 1:} \textit{if a model is confident about its prediction, it should be accurate on the same}.
\item \textbf{Desideratum 2:} \textit{if a model is not confident about its prediction, it may or may not be accurate}.
\end{enumerate}
\vspace{-1.5mm}
This hints at an inverse relation between model accuracy and uncertainty, a property which we exploit when designing metrics for performance evaluation.

There has been a lot of research on deep architectures for semantic segmentation \cite{fcn, unet, segnet, chen2014semantic, chen2017rethinking, chen2018deeplab, deeplab-v3+, aspp2}. In this work, we choose DeepLab-v3+ \cite{deeplab-v3+}, one of the state-of-the-art neural networks for this purpose and create two probabilistic versions of it using dropout based approximate inference techniques: MC dropout \cite{galdropout} and Concrete dropout \cite{concretedropout}. We train these models on the well-known Cityscapes dataset \cite{cityscapes} which contains many images of urban street scenes. Finally, we evaluate and compare the trained models using the metrics which we propose in this work. In Figure \ref{fig:overview}, we present a high level overview of the evaluation system which we implement.

In a nutshell, the main contributions of this paper are:
\vspace{-1.5mm}
\begin{enumerate}[leftmargin=*]
\itemsep-0.3em
\item We propose three novel metrics which can be used to evaluate Bayesian models for semantic segmentation.
\item We create two probabilistic versions of the DeepLab-v3+ \cite{deeplab-v3+} network, which can produce pixel-wise uncertainty estimates in addition to semantic segmentation results.
\item Finally, we evaluate the MC dropout \cite{galdropout} and Concrete dropout \cite{concretedropout} inference techniques using the metrics mentioned above, thereby laying down benchmarks against which other models can be compared. 
\end{enumerate}
\vspace{-1.5mm}
\vspace{-1.5mm}
\section{Related Work} \label{sec:relatedwork}
In this section, we discuss some of the recent works on semantic segmentation as well as those on approximate inference in Bayesian Deep Learning.

\subsection{Semantic Segmentation} \label{subsec:semantic}
The work by Long et al.\ \cite{fcn} was the first of its kind where a convolutional neural network without any fully connected layers was trained in an end-to-end manner directly mapping images to their corresponding segmentation results. This enabled the network to segment images of varying sizes. However, the fully convolutional networks still suffered due to the presence of pooling layers which ignore the positional information of objects in an attempt to reduce the dimensions of feature maps.

In order to get around this issue, researchers have followed two primary threads of thought: the encoder-decoder architecture \cite{unet, segnet, bayesiansegnet} and the dilated/atrous convolutions \cite{dilated, chen2014semantic, chen2017rethinking, chen2018deeplab}. The encoder-decoder networks first reduce the spatial dimensions of the feature maps with repeated applications of convolution and pooling layers in the encoder module. Next, in the decoder module, the spatial dimensions are gradually recovered using de-convolution and upsampling layers. In order to have sharper segmentation results, skip connections are often introduced between the encoder and decoder modules. Few popular works in this category include U-Net \cite{unet}, SegNet \cite{segnet} and RefineNet \cite{refinenet}.

The second class of architectures use dilated or atrous convolutions \cite{dilated} to have a larger field of view over the input feature maps without a decrease in spatial dimensions. One of the most popular set of deep neural networks which follow this policy  is DeepLab \cite{chen2014semantic, chen2017rethinking, chen2018deeplab, deeplab-v3+}. In this work, we adopt a combination of both policies and use DeepLab-v3+ \cite{deeplab-v3+} as the base network for semantic segmentation. DeepLab-v3+ uses atrous convolution layers as well as a simple decoder module to have fine-grained segmentation.

\vspace{-2.0mm}
\subsection{Approximate Inference in Deep Neural Nets} \label{subsec:approxinfer}

The idea behind Bayesian modelling is to find the probability of each set of model parameters given a dataset. In order to do so, an initial distribution known as the \textit{prior} is assumed over the model parameters. Next, with the input of data, this distribution is updated to capture parameters which are more likely to have generated the dataset. This update is done by applying Bayes' theorem. Once the entire dataset is processed, the distribution over the model parameters, known as the \textit{posterior}, captures the updated belief about the optimal set of parameters to represent the data.

However, getting the posterior for large neural networks is computationally intractable. Hence, many methods to approximate the posterior have been proposed. One such class of methods include Markov Chain Monte Carlo (MCMC) techniques \cite{mcmc1, mcmc2, mcmc3}. Another popular set of techniques uses variational inference \cite{var1, bayesbybackprop, galdropout, concretedropout, var2} where the posterior is approximated using a variational distribution. In this work, we have used MC dropout \cite{galdropout} and Concrete dropout \cite{concretedropout} as methods of approximate inference. Both these methods are based on the variational inference approach and are standard well-performing techniques which are easy to implement in deep neural networks.

\subsection{Existing metrics for evaluating uncertainty}
There is a thread of work which focuses on generating calibrated probability estimates from a deep neural network \cite{calibration1, calibration2} as a measure of model confidence. There also exist popular metrics like the expected calibration error (ECE) and the maximum calibration error (MCE) which can be used to quantitatively measure model calibration. However, these metrics are based on softmax probabilities which cannot capture epistemic or model uncertainty \cite{yarinthesis}. Furthermore, there are very simple post-processing techniques like temperature scaling \cite{calibration1} which can make a deterministic and a probabilistic model equally calibrated, thereby rendering the ECE and MCE metrics unable to detect whether a model is "guessing at random", a property which our metrics are designed to capture. We empirically validate these observations in Section \ref{sec:experiment}.

Finally, we conclude this section with a discussion of Bayesian SegNet \cite{bayesiansegnet}, which is one of several works to have applied approximate Bayesian inference in the context of semantic segmentation. The authors of \cite{bayesiansegnet} modify the SegNet architecture \cite{segnet} using MC dropout \cite{galdropout} to obtain uncertainties in addition to segmentation results. Furthermore, they present a few accuracy-vs-uncertainty plots in their work, which are good sanity checks for a Bayesian neural network. These sanity checks are qualitative though, and do not allow us to compare and choose between BDL models for semantic segmentation. It is precisely this gap that we fill by developing \textit{quantitative} measures as well.

\section{Bayesian DeepLab}

In this section we describe the variant of the DeepLab-v3+ \cite{deeplab-v3+} network architecture which we have implemented.

\subsection{Brief review of DeepLab-v3+}
DeepLab-v3+ is one of the state-of-the-art deep neural networks designed for the problem of semantic segmentation. There have been multiple versions of DeepLab, namely DeepLab-v1 and v2 \cite{chen2014semantic, chen2018deeplab}, DeepLab-v3 \cite{chen2017rethinking} and DeepLab-v3+ \cite{deeplab-v3+}, over the years. However, all these versions possess certain common architectural traits. Firstly, they propose atrous or dilated convolutions \cite{atrous1, atrous2, atrous3} as a way to widen the field of view over the input feature maps without increasing the number of parameters or using pooling layers. Secondly, they deal with the problem of objects present at different scales in the image using methods like image pyramid \cite{imagepyramid1, imagepyramid2, imagepyramid3, imagepyramid4}, atrous spatial pyramid pooling (ASPP) \cite{chen2018deeplab, aspp2}, cascaded atrous modules \cite{chen2017rethinking, cascade2, cascade3} and encoder-decoder architectures \cite{segnet, unet}. Thirdly, even though there are no pooling layers in the network, due to the presence of multiple convolution layers with strides of 1 or more, the resulting output dimensions are reduced. In order to regain the original dimensions, the output is passed through a fully connected CRF \cite{chen2014semantic, chen2018deeplab} or resized using bilinear interpolation \cite{chen2017rethinking} or passed through a decoder with learnable parameters \cite{deeplab-v3+}.

Finally, the above-mentioned architectural features can be applied to any base network as long as it is fully convolutional. Some of the popular CNNs which have been used for DeepLab are VGG-16 \cite{vgg}, ResNet-101 \cite{resnet} and Xception \cite{xception}. In this work, we implement DeepLab-v3+ using Xception as the base network. The Xception architecture enjoys the simplicity of VGG with multiple convolution layers stacked on top of one another. Furthermore, Xception modules use skip connections similar to ResNet and are also based on the Inception \cite{inception} hypothesis which postulates the separation of convolution operations performed on spatial (height and width) dimensions and those on the depthwise (or cross-channel) dimensions.

\subsection{Approximate inference in DeepLab-v3+}

In Section \ref{subsec:approxinfer} we listed some techniques for approximate inference in deep neural networks. The purpose of this work is to develop and present metrics with which these inference techniques can be evaluated and benchmarked for the task of semantic segmentation. To do this, we have to first create a probabilistic deep neural network for semantic segmentation. As mentioned before, we use DeepLab-v3+ as the network of choice. Furthermore, we use MC dropout \cite{galdropout} as the primary method of approximate inference. It is one of the current standard baselines and has already been applied to multiple problems in computer vision \cite{bayesiansegnet, multitaskyarin, uncertaintiesforcomputervision}. 

MC dropout works on the principle of variational inference. The idea is to get a posterior distribution $p(\bm{\mathrm{W}}|\bm{\mathrm{X}}, \bm{\mathrm{Y}})$ over the weights $\bm{\mathrm{W}}$ of the neural network, given the training samples $\bm{\mathrm{X}}$ and the corresponding labels $\bm{\mathrm{Y}}$. However, the posterior is intractable and hence, an approximation to the posterior $q(\bm{\textrm{W}})$ known as the variational distribution is defined and the Kullback-Leibler (KL) divergence:
\begin{equation}\label{eq:KLdiv}
    \mathrm{KL}(q(\bm{\textrm{W}}) || p(\bm{\mathrm{W}}|\bm{\mathrm{X}}, \bm{\mathrm{Y}}))
\end{equation}
between the actual posterior and the variational distribution is minimized. In MC dropout, the variational distribution defined over the network weights is a Bernoulli distribution. In \cite{galdropout}, the authors observed that placing a Bernoulli distribution with parameter $p_b$ over the weights of a hidden layer is equivalent to performing dropout on that layer with a dropout rate of $p_b$. Furthermore, they also noted that minimising the well-known cross-entropy loss function using standard optimisation algorithms like stochastic gradient descent has the desired effect of minimizing the KL divergence term in equation \ref{eq:KLdiv}. Hence, in order to perform approximate inference, one first needs to train a network with dropout. However, unlike common practice, these dropout layers are kept active even during the test phase. The idea is to get samples from the posterior distribution and as the dropout layers place a Bernoulli distribution over the network weights, performing a stochastic forward pass through a trained network can be interpreted as generating a Monte Carlo sample from the posterior distribution. Therefore, multiple forward passes on the same input generate multiple such Monte Carlo samples, the mean of which can then be used as the network prediction and the variance can be interpreted as an uncertainty estimate.

Ideally, in a Bayesian neural network, a dropout layer should be inserted after every hidden layer of the network. However, as was observed by Kendall et al. in \cite{bayesiansegnet} for the SegNet architecture and as we observe for DeepLab-v3+, insertion of dropout layers after every convolution layer in a large neural network regularises it to an extent that makes training prohibitively slow. Therefore, the dropout layers have to be inserted only in certain regions of the network. This gives rise to multiple probabilistic variants of the Bayesian DeepLab architecture depending on where in the network, the dropout layers are inserted. However, for the sake of simplicity, in this work, we insert dropout layers only in the middle flow of the DeepLab-v3+ network. We do this because of the hypothesis proposed in \cite{bayesiansegnet} which states that low level features in shallower layers of the network are mostly consistent across the distribution of models and hence can be represented using deterministic weights whereas the higher level features in deeper layers are better modelled using probabilistic weights.

With the incorporation of MC dropout in DeepLab-v3+, we end up with Bayesian DeepLab, a probabilistic deep neural network designed specifically for semantic segmentation. However, in this work, our goal is to develop evaluation metrics for such networks and lay out a few benchmarks based on these metrics. Thus, in order to compare performance with MC dropout, we implement an alternative inference technique, namely Concrete dropout \cite{concretedropout}. We choose Concrete dropout in particular because both the approximate inference techniques are dropout based methods. They are also simple to implement and require minimal changes to the network architecture. Concrete Dropout is a modification on the MC dropout method where the network tunes the dropout rates during training. Similar to the MC dropout variant, we place the concrete dropout layers in the middle flow of the DeepLab-v3+ network.

\subsection{Network Architecture}

The backbone framework of Bayesian DeepLab is similar to DeepLab-v3+ \cite{deeplab-v3+} where the inputs are first passed through an extended Xception \cite{xception} network followed by an ASPP module for multi-scale image processing and finally a decoder module to resize the images to the original input dimensions and to produce sharp segmentation results. The differences between the network architecture which we use in this work and the original DeepLab-v3+ network proposed in \cite{deeplab-v3+} are as follows:
\vspace{-1.80mm}
\begin{enumerate}[leftmargin=*]
\itemsep-0.3em
\item Bayesian DeepLab has dropout layers at different points in the network. In particular we insert a dropout layer after every 4 Xception modules in the middle flow of the network. As there are 16 Xception modules in the middle flow, there are a total of 4 dropout layers in the network. We use a dropout rate of 0.5 in each of the dropout layers. However, these rates are hyperparameters which can be fine-tuned further.
\item We do not use cascaded atrous modules or image pyramids. The only method for multi-scale image processing adopted by Bayesian DeepLab is Atrous Spatial Pyramid Pooling (ASPP). We do this primarily for the sake of simplicity and to reduce training time.
\end{enumerate}
\vspace{-1.80mm}
We present the Bayesian DeepLab architecture in Figure \ref{fig:bayesian_deeplab_dropout} in the appendix.

\subsection{Uncertainty Metrics}

There are two types of uncertainties which we study in this work. \textit{Epistemic uncertainty}, also known as \textit{model uncertainty} represents what the model does not know due to insufficient training data. This kind of uncertainty can be explained away with more training data. \textit{Aleatoric uncertainty} is caused due to noisy measurements in the data and can be explained away with increased sensor precision (but cannot be explained away with increase in training data). The two uncertainties combined form the \textit{predictive uncertainty} of the network. In \cite{yarinthesis}, the author suggests some information theoretic metrics which can be used as measures of uncertainty in classification problems. In our work, we use two such metrics, namely the \textit{entropy} of the predictive distribution (also known as \textit{predictive entropy}) and the \textit{mutual information} between the predictive distribution and the posterior over network weights.

The predictive entropy $\mathbb{\hat{H}}[y|\bm{\mathrm{x}}, \mathcal{D}_{train}]$ given a test input $\bm{\mathrm{x}}$ and the training data $\mathcal{D}_{train}$ can be approximated as:
\begin{dmath}
    \mathbb{\hat{H}}[y|\bm{\mathrm{x}}, \mathcal{D}_{train}] = - \sum_{c}\left(\frac{1}{T} \sum_{t}p(y=c|\bm{\mathrm{x}}, \hat{w}_{t})\right)\log\left(\frac{1}{T} \sum_{t}p(y=c|\bm{\mathrm{x}}, \hat{w}_{t})\right)
\end{dmath}
where $y$ is the output variable, $c$ ranges over all the classes, $T$ is the number of Monte Carlo samples (stochastic forward passes), $p(y=c|\bm{\mathrm{x}}, \hat{w}_{t})$ is the softmax probability of input $\bm{\mathrm{x}}$ being in class $c$, and $\hat{w}_{t}$ are the model parameters on the $t^{th}$ Monte Carlo sample. Similarly, the mutual information between the predictive distribution and the posterior over model parameters can be approximated as:
\begin{multline}
    \mathbb{\hat{I}}[y, w|\bm{\mathrm{x}}, \mathcal{D}_{train}] = \mathbb{\hat{H}}[y|\bm{\mathrm{x}}, \mathcal{D}_{train}] \\ + \frac{1}{T}\sum_{c,t}p(y=c|\bm{\mathrm{x}}, \hat{w}_{t})\log p(y=c|\bm{\mathrm{x}}, \hat{w}_{t}).
\end{multline}
We choose the predictive entropy and mutual information metrics as they capture different kinds of uncertainty. As observed in \cite{yarinthesis}, mutual information captures epistemic or model uncertainty whereas predictive entropy captures predictive uncertainty which combines both epistemic and aleatoric uncertainties. It is worth noting here that in semantic segmentation, we produce pixel-wise classification results and hence, we also produce pixel-wise uncertainty estimates. Thus, we get uncertainty maps which have the same dimensions as that of the input image.

\section{Performance Evaluation Metrics} \label{sec:evalmetrics}

\begin{figure}[t]
\begin{center}
   \includegraphics[width=\linewidth]{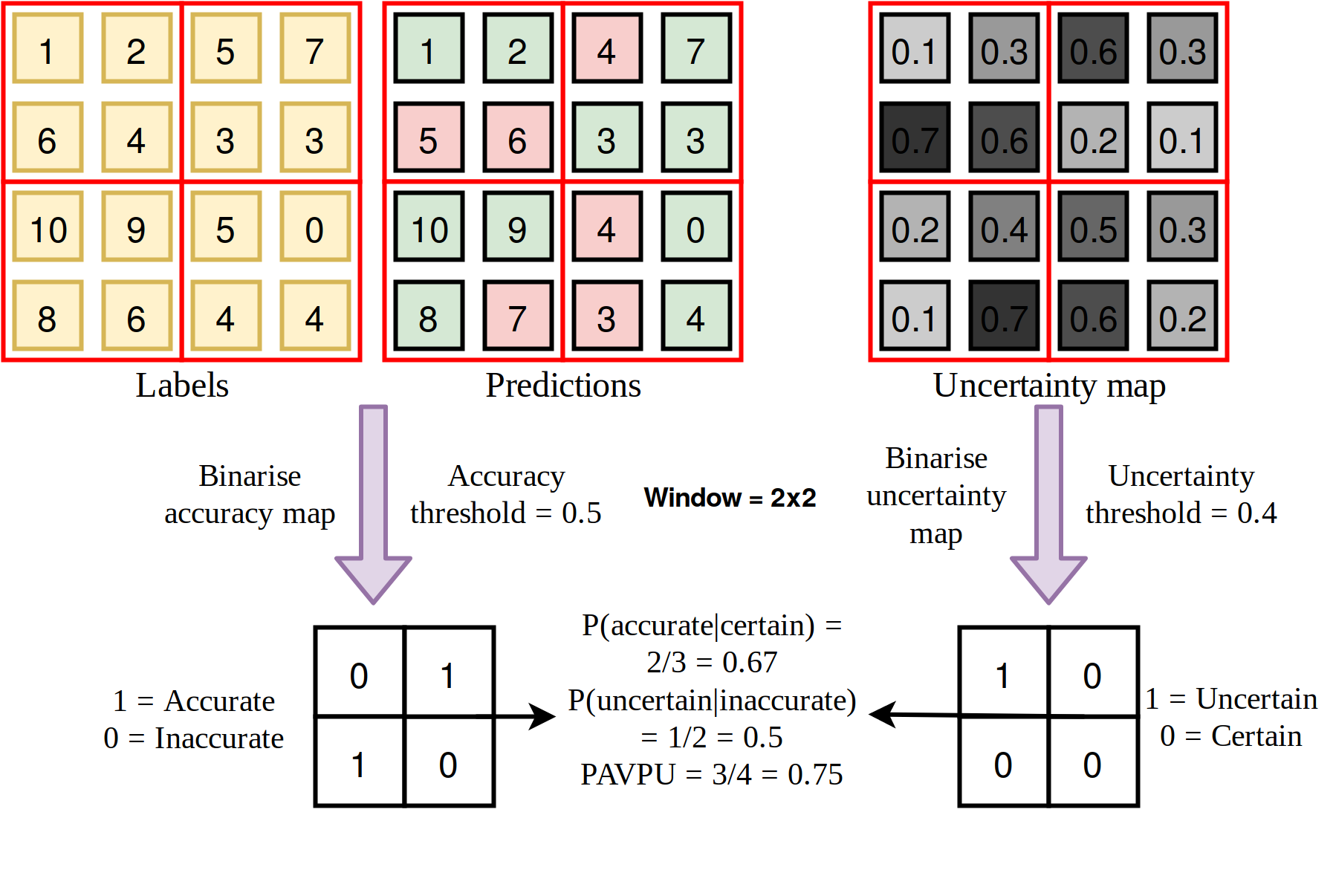}
\end{center}
\vspace{-1\baselineskip}
   \caption{Worked out example of computing the performance evaluation metrics for Bayesian models in semantic segmentation.}
\vspace{-1\baselineskip}
\label{fig:metrics}
\end{figure}

As mentioned in Section \ref{sec:intro}, we measure the performance of Bayesian models using metrics which capture properties which we want the model to satisfy. In particular, we assume that \textit{if a model is confident about its prediction, it should be accurate on the same.} This also implies that \textit{if a model is inaccurate on an output, it should be uncertain about the same output.} It is worth noting that the converse of these assumptions may not hold. For instance, a model may have a high epistemic uncertainty on a class which appears infrequently in the training set but can still be accurate on its prediction. Given the above assumptions, we can define the following two conditional probabilities:
\vspace{-1.50mm}
\begin{enumerate}[leftmargin=*]
    \itemsep-0.3em
    \item $p(\bm{\mathrm{accurate}}|\bm{\mathrm{certain}})$: The probability that the model is accurate on its output given that it is confident on the same.
    \item $p(\bm{\mathrm{uncertain}}|\bm{\mathrm{inaccurate}})$: The probability that the model is uncertain about its output given that it has made a mistake in its prediction (i.e., is inaccurate).
\end{enumerate}
\vspace{-1.50mm}
In order to implement the above metrics, we first choose a patch/window size $w$ and traverse the predicted labels, actual labels and uncertainty maps using windows of dimensions $w \times w$, much like the traversal in a convolution operation. Technically, the patch dimensions can be as small as a single pixel (i.e., a $1 \times 1$ patch) or as large as the entire image. However, labelling the entire image as accurate or uncertain is not very useful. Furthermore, the uncertainties occur in regions within the image comprising multiple pixels in close neighbourhoods. Capturing these uncertain regions is useful for downstream tasks and having $1 \times 1$ single pixel patches does not help in this regard. Thus, we have $w > 1$ and we compute the accuracy of each patch from the predicted and actual labels. This accuracy can be computed using any standard technique. In this work we use the \textit{pixel accuracy} metric defined in \cite{fcn}. If the patch accuracy is above a certain threshold, we mark the patch as \textit{accurate}.

\begin{figure*}[t]
\begin{center}
   \includegraphics[width=\textwidth]{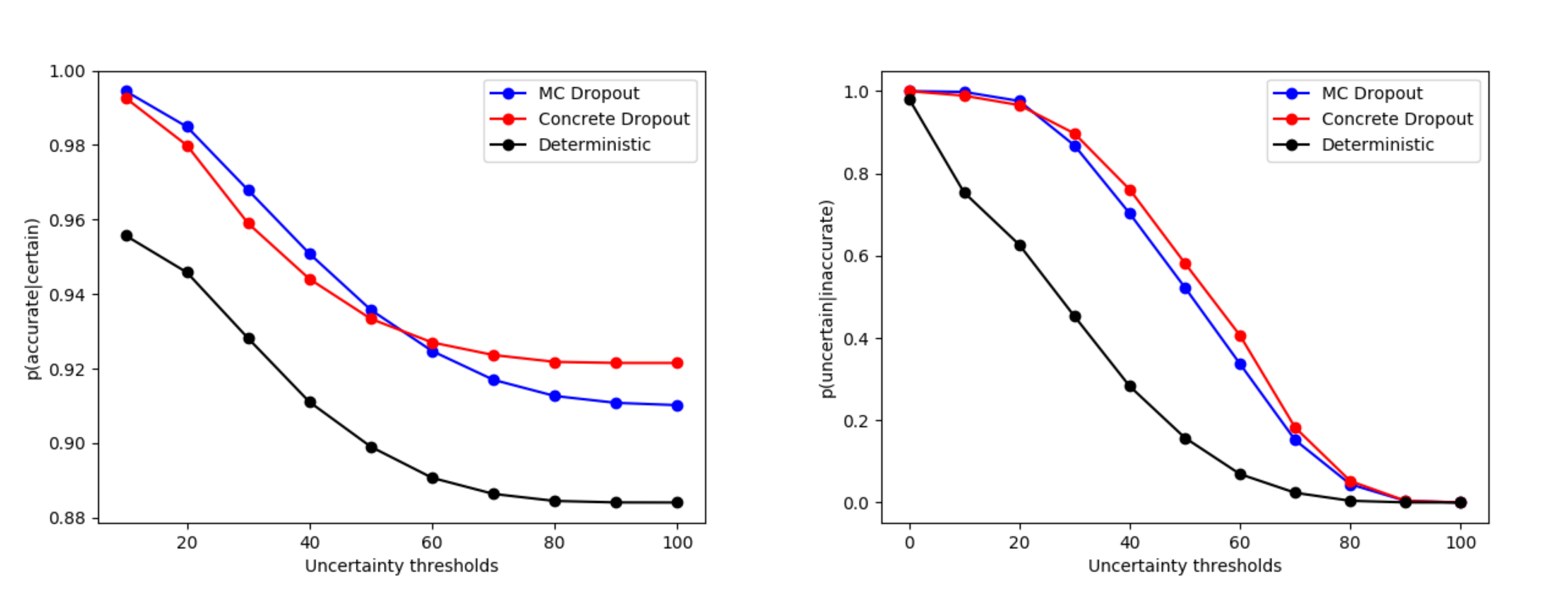}
\end{center}
\vspace{-.5\baselineskip}
   \caption{\textbf{Plots of p(accurate|certain) and p(uncertain|inaccurate) for varying thresholds of uncertainty.}}
\vspace{-1.2\baselineskip}
\label{fig:evalmets1}
\end{figure*}

\begin{figure}[t]
\vspace{-1.5mm}
\begin{center}
   \includegraphics[width=\linewidth]{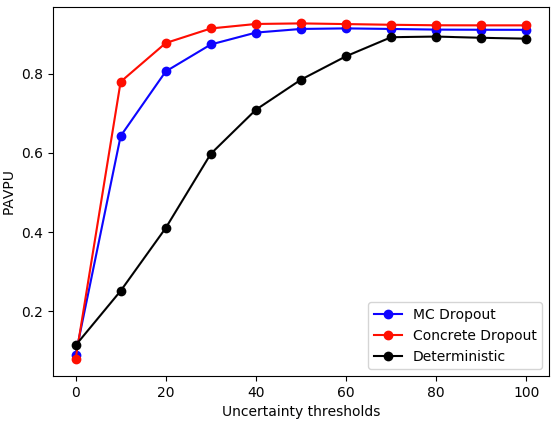}
\end{center}
\vspace{-.5\baselineskip}
   \caption{\textbf{Plot of PAvPU for varying thresholds of uncertainty.}}
\vspace{-1.2\baselineskip}
\label{fig:evalmets2}
\end{figure}

Similarly, from the corresponding patch obtained from the uncertainty map, we compute the average patch uncertainty. If this uncertainty value is above a given threshold, we label the patch as \textit{uncertain}. There can be multiple ways of setting the uncertainty threshold. One simple way would be to find the average uncertainty of all pixels over a validation set and use that value as the threshold. Other ways could include computing the minimum $u_{min}$ and maximum $u_{max}$ uncertainty values over validation set pixels and setting the uncertainty threshold $u_{th}$ as:
\begin{equation}
    u_{th} = u_{min} + (t(u_{max} - u_{min})).
\end{equation}
where $t$ is in $[0, 1]$. Once the entire dimensions of the image have been covered, we construct a confusion matrix containing the number of patches which are accurate and certain ($n_{ac}$), accurate and uncertain ($n_{au}$), inaccurate and certain ($n_{ic}$) and inaccurate and uncertain ($n_{iu}$). We can then report the conditional probabilities $p(\bm{\mathrm{accurate}}|\bm{\mathrm{certain}})$ and $p(\bm{\mathrm{uncertain}}|\bm{\mathrm{inaccurate}})$ as follows:
\begin{equation}\label{eq:acccert}
    p(\bm{\mathrm{accurate}}|\bm{\mathrm{certain}}) = \frac{n_{ac}}{(n_{ac} + n_{ic})}
\end{equation}
\begin{equation}\label{eq:uncinacc}
    p(\bm{\mathrm{uncertain}}|\bm{\mathrm{inaccurate}}) = \frac{n_{iu}}{(n_{ic} + n_{iu})}
\end{equation}
Finally, we combine both the good cases of (accurate, certain) and (inaccurate, uncertain) patches into a single metric, the \textit{Patch Accuracy vs Patch Uncertainty} (PAvPU), defined as follows:
\begin{equation}\label{eq:pavpu}
    \mathrm{PAvPU} = \frac{(n_{ac} + n_{iu})}{(n_{ac} + n_{au} + n_{ic} + n_{iu})}
\end{equation}
Clearly, a model with a higher value of the above metrics is a better performer. As the values of the above metrics depend on three parameters: the accuracy threshold, the uncertainty threshold, and the patch dimensions, an interesting experiment would be to observe how the metrics vary with these parameters. We show this in Figures \ref{fig:evalmets1} and \ref{fig:evalmets2}. In Figure \ref{fig:metrics}, we provide an illustrative example of computing the above three metrics.

\section{Experiments and Results} \label{sec:experiment}
In this section we first evaluate the trained models on their segmentation performance using the pixel accuracy, mean accuracy and mean IOU metrics defined in \cite{fcn}. Next, we compare and benchmark MC dropout \cite{galdropout} and Concrete dropout \cite{concretedropout} inference using the metrics proposed in Section \ref{sec:evalmetrics}. We also compare the above two Bayesian models with a deterministic DeepLab-v3+ baseline where we use the entropy of the softmax distribution as a measure of uncertainty. All experiments have been performed on Cityscapes \cite{cityscapes} dataset. We provide further details about the training infrastructure in Appendix \ref{app:traininginfra}.

\begin{figure*}[t]
\begin{center}
   \includegraphics[width=\textwidth]{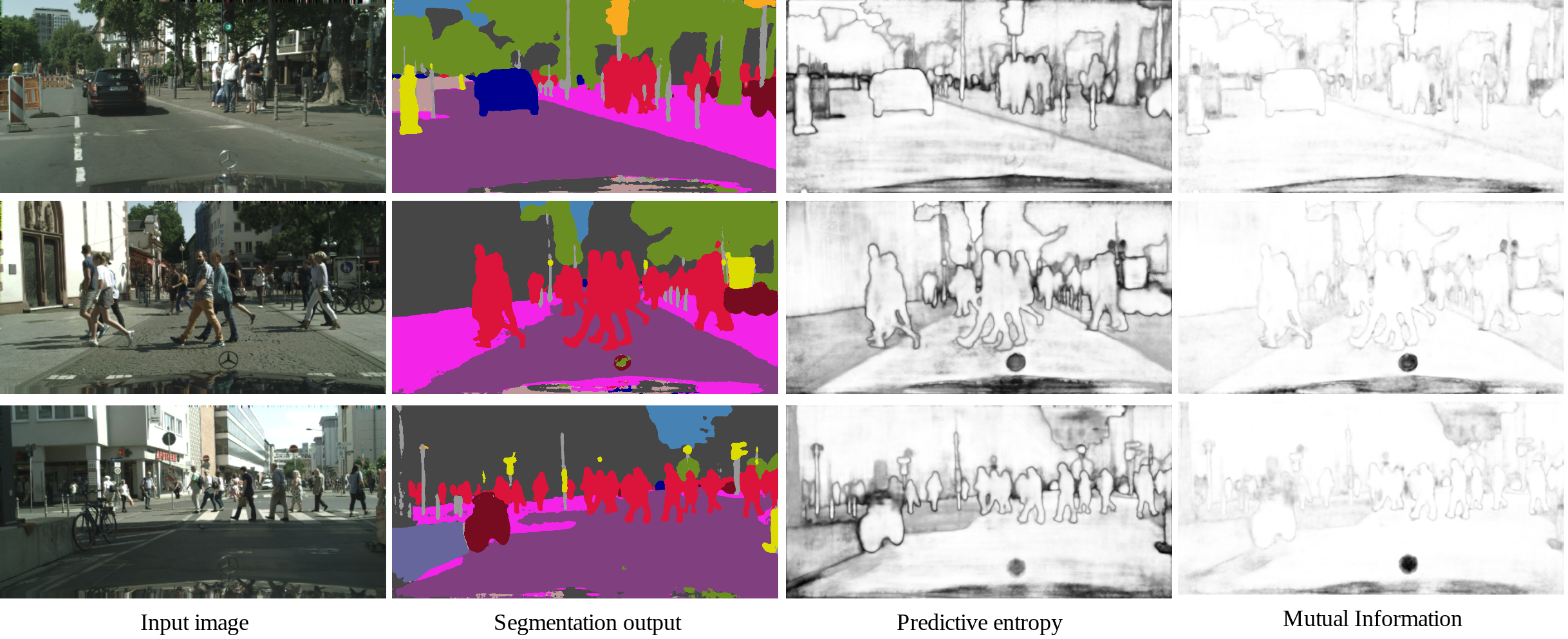}
\end{center}
   \caption{\textbf{Qualitative results for semantic segmentation with uncertainty estimates on Cityscapes images.} The results include images from the Cityscapes \textit{val} set, the corresponding semantic segmentation results from our model and the predictive and epistemic uncertainties estimated through the predictive entropy and the mutual information metrics respectively. Darker shades indicate higher uncertainty.}
\vspace{-.5\baselineskip}
\label{fig:qualitative_result}
\end{figure*}

\subsection{Semantic Segmentation Performance}
In Table \ref{table:segresults} we report the semantic segmentation results obtained from the Bayesian DeepLab variants using both MC dropout and Concrete dropout inference and compare these results with different versions of DeepLab \cite{chen2018deeplab, chen2017rethinking, deeplab-v3+} on the Cityscapes \textit{val} set. We observe that Concrete dropout performs better than MC dropout with respect to all the three metrics: pixel accuracy, mean accuracy and mean IOU. Furthermore, both our models with mean IOU values of 78.05 and 79.12 outperform all the DeepLab versions except DeepLab-v3+ which has a mean IOU of 79.14. It is worth noting that we compare our models only with those versions of DeepLab which use the same hyperparameters as us. To be precise, we compare with the version of DeepLab-v3 which uses an output stride of 16 and DeepLab-v3+ which uses Xception-65 with ASPP and decoder modules but without image level features.

In Figure \ref{fig:qualitative_result}, we present some qualitative results on Cityscapes \textit{val} set images. It is interesting to note the difference between the uncertainty maps provided by the predictive entropy and mutual information metrics. In case of mutual information, we observe high uncertainty inside the boundaries of objects which the model is confused about. The bonnet of the Mercedes at the bottom of each image is one such example. However, in predictive entropy maps, we also see high uncertainties on the edges of objects like pedestrians or cars. These edges are regions where the presence of noise in the dataset is highly likely. This observation supports the explanation in \cite{yarinthesis} that mutual information captures epistemic or model uncertainty and predictive entropy captures aleatoric uncertainty.

\subsection{Evaluation of Bayesian DeepLab}

\begin{table}
\begin{center}
\begin{tabular}{c|c|c|c}
\toprule
\textbf{Method} & \begin{tabular}{@{}c@{}}\textbf{Pixel} \\ \textbf{Accuracy} \end{tabular} &  \begin{tabular}{@{}c@{}}\textbf{Mean} \\ \textbf{Accuracy} \end{tabular} & \begin{tabular}{@{}c@{}}\textbf{Mean} \\ \textbf{IOU} \end{tabular} \\
\bottomrule
\multicolumn{4}{c}{\textit{DeepLab}} \\
\bottomrule
\begin{tabular}{@{}c@{}}DeepLab \\ (VGG-16) \cite{chen2018deeplab} \end{tabular} & NA & NA & 65.94 \\
\hline
\begin{tabular}{@{}c@{}}DeepLab \\ (ResNet-101) \cite{chen2018deeplab} \end{tabular} & NA & NA & 71.40 \\
\hline
\begin{tabular}{@{}c@{}}DeepLab-v3 \\ (OS=16) \cite{chen2017rethinking} \end{tabular} & NA & NA & 77.23 \\
\hline
\begin{tabular}{@{}c@{}}DeepLab-v3+ \\ (X-65) \cite{deeplab-v3+} \end{tabular} & NA & NA & \textbf{79.14} \\
\bottomrule
\multicolumn{4}{c}{\textit{Bayesian DeepLab}} \\
\bottomrule
MC Dropout & 95.31 & 85.11 & 78.05 \\
\hline
Concrete Dropout & \textbf{96.47} & \textbf{87.26} & 79.12 \\
\bottomrule
\end{tabular}
\end{center}
\caption{\textbf{Semantic segmentation performance on Cityscapes \textit{val} set for Bayesian DeepLab networks.} Best results are in bold.}
\vspace{-1\baselineskip}
\label{table:segresults}
\end{table}

In order to compute the metrics $p(\bm{\mathrm{accurate}}|\bm{\mathrm{certain}})$, $p(\bm{\mathrm{uncertain}}|\bm{\mathrm{inaccurate}})$ and PAvPU, we use patches with dimensions $4 \times 4$ although other patch dimensions can also be used. For the sake of simplicity, we set the accuracy threshold to 0.5 (or 50\%) and the uncertainty threshold to the mean uncertainty value in the validation set.

As seen in Table \ref{table:metricvals}, the dropout based models outperform the deterministic model and Concrete dropout consistently outperforms MC dropout on all the metrics. It is worth noting that we cannot compute the metrics for mutual information for a deterministic model as the mutual information is always 0, thereby indicating that the deterministic model cannot capture epistemic uncertainty. In Figures \ref{fig:evalmets1} and \ref{fig:evalmets2}, we plot these metrics for varying thresholds of uncertainty. We can draw the following observations from the plots:
\vspace{-1.5mm}
\begin{enumerate}[leftmargin=*]
\itemsep-0.3em
\item When the uncertainty threshold is 0, all the patches are marked uncertain. Hence $n_{ac} + n_{ic} = 0$ in equation \ref{eq:acccert} and the value of $p(\bm{\mathrm{accurate}}|\bm{\mathrm{certain}})$ is undefined. Hence, for the plot of $p(\bm{\mathrm{accurate}}|\bm{\mathrm{certain}})$ we start the uncertainty threshold from 10\%. Furthermore, $n_{ic} + n_{iu} = n_{iu}$ in equation \ref{eq:uncinacc} and therefore, $p(\bm{\mathrm{uncertain}}|\bm{\mathrm{inaccurate}})$ is 1. Lastly, as $n_{ac} = n_{ic} = 0$, the PAvPU metric (as defined in equation \ref{eq:pavpu}) becomes $\frac{n_{iu}}{n_{iu} + n_{au}}$ which in this case, is the fraction of inaccurate patches in the image.

\item When the uncertainty threshold is 100\%, all the patches are labelled certain. In this case, both $p(\bm{\mathrm{accurate}}|\bm{\mathrm{certain}})$ and PAvPU boil down to the fraction of patches which are accurate. As $n_{iu} = 0$, $p(\bm{\mathrm{uncertain}}|\bm{\mathrm{inaccurate}})$ reduces to 0.

\item It is interesting to note that the sum of the PAvPU values at uncertainty thresholds 0 and 100\% is 1. Furthermore, the PAvPU value at 100\% uncertainty threshold is significantly greater than the value at 0. This indicates that the number of accurate patches is much higher than the number of inaccurate patches.

\item We observe that the Bayesian models perform significantly better than the deterministic baseline. One reason behind this is that the softmax entropy in a deterministic model is only able to capture aleatoric uncertainty and regions of high epistemic uncertainty are ignored, resulting in a relatively high number of inaccurate but certain patches. This does not occur in the Bayesian models as predictive entropy captures both aleatoric and epistemic uncertainties. Finally, the plots indicate the superior performance of Concrete dropout over MC dropout which is consistent with the results in Table \ref{table:metricvals}.
\end{enumerate}
\begin{figure*}[t]
\begin{center}
   \includegraphics[width=\textwidth]{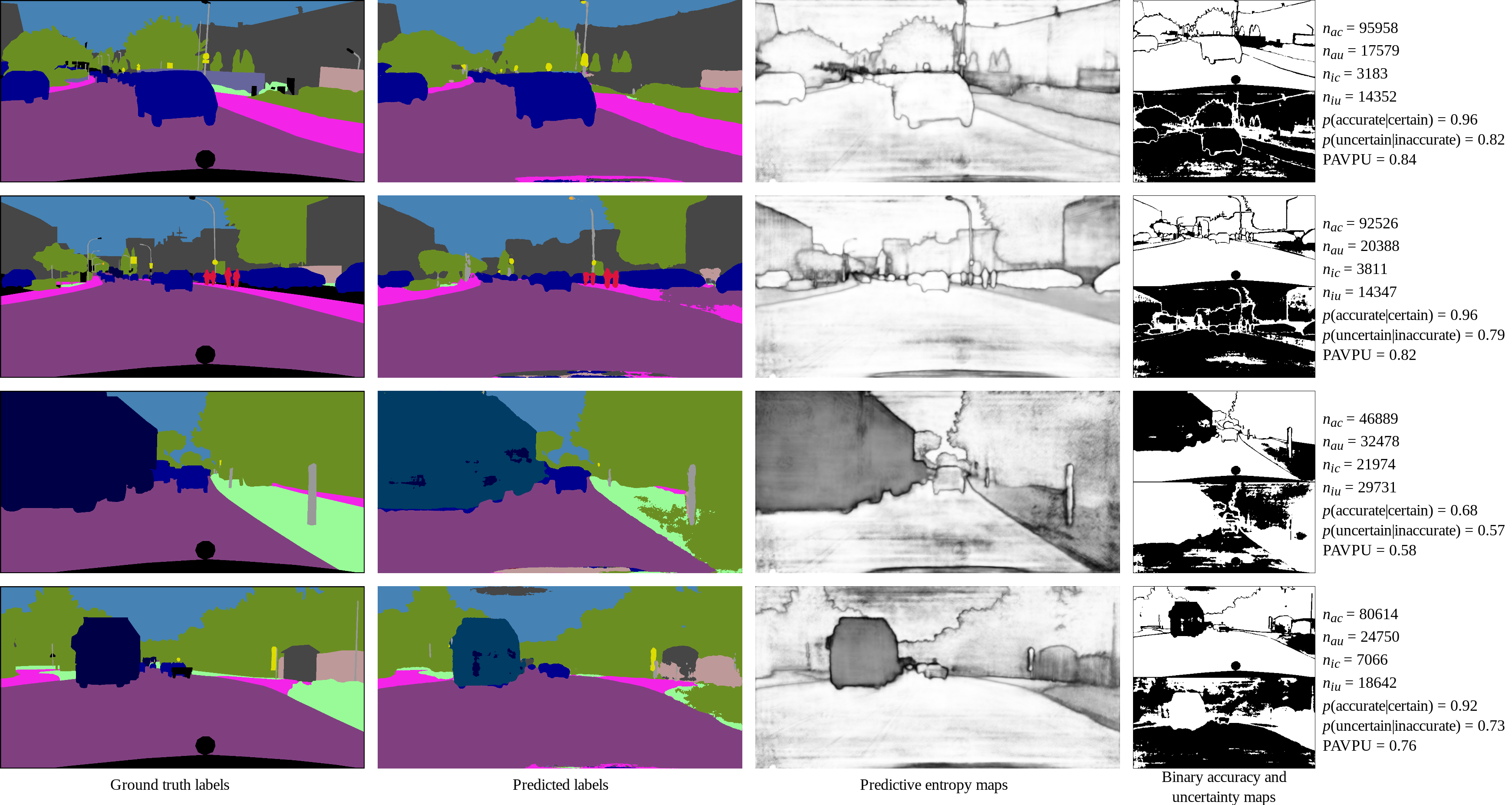}
\end{center}
   \caption{\textbf{Computation of $p(\bm{\mathrm{accurate}}|\bm{\mathrm{certain}})$, $p(\bm{\mathrm{uncertain}}|\bm{\mathrm{inaccurate}})$ and PAvPU on real world images.} We take two success and two failure cases and compute the metrics for each of these cases. The upper binary map in each row is the accuracy map and the lower map corresponds to the uncertainty map. White regions in these accuracy/uncertainty maps represent accurate/uncertain patches.}
\vspace{-1\baselineskip}
\label{fig:analysis}
\end{figure*}
\vspace{-1.5mm}
In Figure \ref{fig:analysis} we show the computation of the $p(\bm{\mathrm{accurate}}|\bm{\mathrm{certain}})$, $p(\bm{\mathrm{uncertain}}|\bm{\mathrm{inaccurate}})$ and PAvPU metrics for real world images using predictive entropy. We have two successful cases where the network makes accurate predictions and two failure cases where the network misclassifies a truck as a bus. We can see from the predictive entropy that the model is uncertain on the misclassified pixels. For every case, we have two binary maps, one for accuracy and one for uncertainty. Each pixel in a binary map corresponds to a patch in the image. A white pixel represents an accurate/uncertain patch and a black pixel represents the opposite. It is interesting to note how these maps show the inverse relation between model accuracy and uncertainty. Furthermore, in the two failure cases, we observe that $n_{au}$ and $n_{ic}$ values (representing the undesirable patches) are relatively high. \textbf{This in turn, gets reflected in the values of the metrics which are lower for the failure cases compared to the successful ones}.

\begin{table}
\begin{center}
\resizebox{\linewidth}{!}{
\begin{tabular}{c|c|c|c}
\toprule
\textbf{Method} & \begin{tabular}{@{}c@{}}\textbf{p(accurate|} \\ \textbf{certain)} \end{tabular} &  \begin{tabular}{@{}c@{}}\textbf{p(uncertain|} \\ \textbf{inaccurate)} \end{tabular} & \textbf{PAvPU} \\
\bottomrule
\multicolumn{4}{c}{\textit{Predictive Entropy}} \\
\bottomrule
Deterministic & 0.9527 & 0.7068 & 0.6562 \\
MC Dropout & 0.9869 & 0.8962 & 0.7861 \\
Concrete Dropout & \textbf{0.9909} & \textbf{0.9144} & \textbf{0.8034} \\
\bottomrule
\multicolumn{4}{c}{\textit{Mutual Information}} \\
\bottomrule
MC Dropout & 0.9630 & 0.6720 & 0.8267 \\
Concrete Dropout & \textbf{0.9669} & \textbf{0.7074} & \textbf{0.8530} \\
\bottomrule
\end{tabular}}
\end{center}
\caption{Performance of Bayesian DeepLab variants and a deterministic DeepLab network evaluated using the three metrics proposed in Section \ref{sec:evalmetrics}. Best results are shown in bold.}
\vspace{-1\baselineskip}
\label{table:metricvals}
\end{table}

As mentioned in section \ref{subsec:approxinfer}, we compare the ECE and MCE \cite{calibration2} values of the three models after calibrating them using temperature scaling \cite{calibration1}. With 15 bins, we obtain an ECE of 0.0182, 0.0161 and 0.0165 at optimal temperatures 4.7, 3.1 and 3.2 respectively for the deterministic, MC dropout and Concrete dropout models. Similarly, we obtain MCE values of 0.1969, 0.1823 and 0.1785 at the same optimal temperatures. Thus, after temperature scaling, the ECE and MCE metrics are non-indicative of which model to choose and are not able to detect when a model is guessing at random (i.e., has high epistemic uncertainty). This, however is clearly captured in the proposed metrics as can be seen from Figures \ref{fig:evalmets1} and \ref{fig:evalmets2} and Table \ref{table:metricvals}. Finally, it is worth noting that the proposed metrics evaluate uncertainty values and hence, should be used in addition to conventional metrics of measuring accuracy like mean IOU \cite{fcn}.

\vspace{-1.5mm}
\section{Conclusions}

In this work we have developed metrics to evaluate Bayesian models for the task of semantic segmentation. We have created two probabilistic variants of the DeepLab-v3+ \cite{deeplab-v3+} network and have evaluated them using these metrics, thereby providing benchmarks which can be used for future comparisons. For experiments, we have used the Cityscapes \cite{cityscapes} dataset which is particularly suited for applications like autonomous driving. An interesting future work would be to develop metrics which measure performance of Bayesian models based on how effective the segmentation outputs and uncertainty estimates are in making safe and correct autonomous driving decisions. This idea can be further extended to include other downstream applications as well where semantic segmentation is a useful intermediate tool.

{\small
\bibliographystyle{ieee}
\bibliography{egbib}
}

\newpage
\clearpage

\onecolumn
\begin{appendices}

\section{Bayesian DeepLab Network Architecture} \label{app:bayesdeeplab}

\begin{figure}[h]
\centering
   \includegraphics[width=\linewidth]{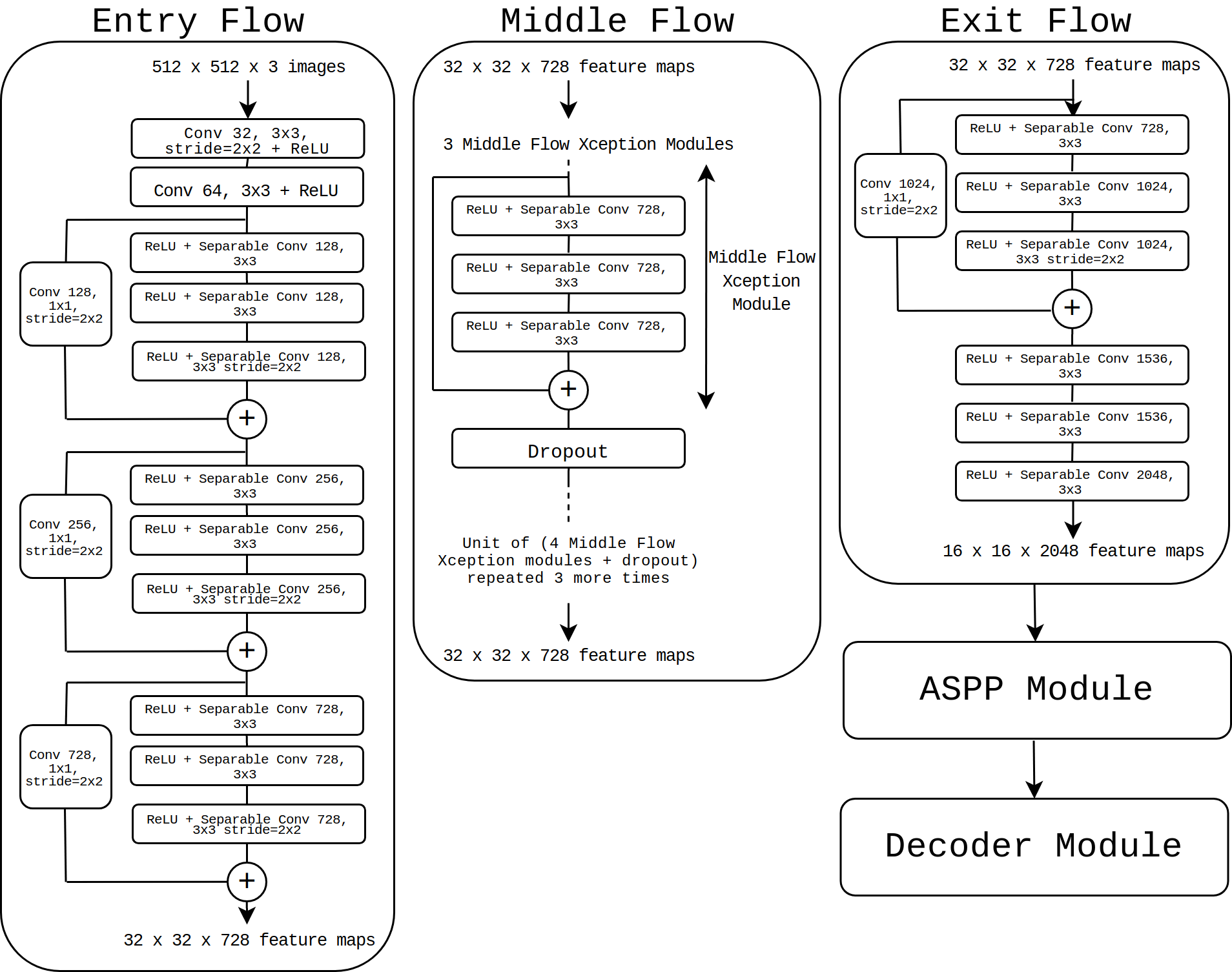}
   \caption{\textbf{Bayesian DeepLab network architecture: }The inputs are passed in order first through the entry flow followed by the middle flow and finally through the exit flow. The ASPP module is used to recognize objects at different scales and the Decoder module resizes the outputs to the original input dimensions.}
\label{fig:bayesian_deeplab_dropout}
\end{figure}

In Figure \ref{fig:bayesian_deeplab_dropout}, we present the network architecture of Bayesian DeepLab using MC dropout inference. We use Xception as the base network. However, there are certain differences between the original Xception \cite{xception} architecture and the one which we use. The differences are as follows:
\begin{enumerate}
    \item There are no pooling layers in our network. We use separable convolution filters with a stride of 2 instead of max pooling layers. This helps in dense predictions. Furthermore, following the FCN \cite{fcn} philosophy, our network is fully convolutional and hence can segment images of arbitrary sizes.
    \item The middle flow in our network has 16 modules instead of 8 as described in the original Xception paper \cite{xception}.
\end{enumerate}

\newpage
\section{Training Infrastructure} \label{app:traininginfra}
We train all our networks on the Cityscapes \cite{cityscapes} dataset, one of the most popular datasets for urban scene understanding. It has 5000 images collected from street scenes in 50 different cities. There are 2975 images in the training set, 500 images in the validation set and 1525 test images with dimensions $2048 \times 1024$. In order to train the networks, we set the following parameters:
\vspace{-1.5mm}
\begin{enumerate}[leftmargin=*]
\itemsep-0.3em
\item a list of \textit{atrous rates} for the ASPP module which we set to $[6, 12, 18]$ following the DeepLab-v3+ \cite{deeplab-v3+} paper,
\item the \textit{output stride} and the \textit{decoder output stride} \cite{chen2017rethinking} which we set as 16 and 4 respectively,
\item the \textit{crop size} for training images which we set to $512 \times 512$ and
\item the \textit{training batch size} which we set to 16.
\end{enumerate}
\vspace{-1.5mm}
Furthermore, the Concrete dropout layers require two additional hyperparameters: the \textit{weight regulariser} which is set to $1e-8$ and the \textit{dropout regulariser} which we set to $1/(n \times h \times w)$, where $n$ is the number of training images and $h$ and $w$ are the height and width of the image respectively, following the paper \cite{concretedropout}. We train all the networks for 90,000 iterations on 8 NVIDIA Tesla P100-SXM2 GPUs. The networks finish training in approximately 3 days.

\end{appendices}
\end{document}